\documentclass{statsoc}

\usepackage{mathsym}

\newcommand{\bx}{{\bf x}}

\newcommand{\cP}{{\cal P}}

\newcommand{\cX}{{\cal X}}
\newcommand{\cY}{{\cal Y}}
\newcommand{\cZ}{{\cal Z}}

\newtheorem{theorem}{Theorem}

\newtheorem{corollary}{Corollary}

\newtheorem{definition}[theorem]{Definition}

\title[Learning Bayesian Networks using Statistical Learning Theory]{Reliable and Efficient Inference of Bayesian Networks \\from Sparse Data
by Statistical Learning Theory}

\author[D. Janzing {\it et al.}]{Dominik Janzing}
\address{Institut f\"{u}r Algorithmen und Kognitive Systeme ,
Universit\"{a}t Karlsruhe \\ Am Fasanengarten 5, 76131 Karlsruhe, Germany, email
 janzing@ira.uka.de}

\author[D. Janzing and D. Herrmann]{Daniel J. L. Herrmann}
\address{Max-Planck-Institut for Biological Cybernetics , Spemannstr. 38,
72 076 T\"{u}bingen , Germany}

\begin{document}

\begin{abstract}
To learn (statistical) dependencies among  random variables
requires exponentially large sample size  in the number of
observed random variables if any arbitrary joint probability
distribution can occur.

We consider the case that sparse data strongly suggest that the
probabilities can be described by a {\it simple} Bayesian network,
i.e., by a graph with small in-degree $\Delta$. Then this simple
law will also explain further data with high confidence. This is
shown by calculating bounds on the VC dimension of the set of
those probability measures that correspond to simple graphs. This
allows to select networks by  structural risk minimization and
gives reliability bounds on the error of the estimated joint
measure without (in contrast to a previous paper) any prior
assumptions on the set of possible joint measures.

The complexity for searching the optimal Bayesian networks
of in-degree $\Delta$ increases only polynomially in
the number of random varibales
for constant $\Delta$
and
the optimal joint measure associated with a given graph
can be found by convex optimization.
\end{abstract}


\section{Bayesian networks and the causal Markov condition}

\label{Intro}

Learning statistical dependencies among a set of $n$ random variables
$X_1,\dots,X_n$ is an important tool of  scientific research.
Formally, the task of learning those dependencies is to obtain
some information about the joint probability measure $P$ where
$P(x_1,\dots,x_n)$ denotes the probability of the event
$X_1=x_1,\dots, X_n=x_n$\footnote{Here we assume that each random
variable $X_j$ can only take values in a finite set $\Omega_j$.} .
A useful way  to represent such information in a graphical way is
given by the concept of {\it Bayesian networks} (Pearl, 1985).

Although one may consider Bayesian networks merely as a way of
encoding statistical dependencies into a graph, the concept
is better understood if a {\it causal} interpretation is
given to the graph. Recall that every joint probability $P$ can be
factorized as
\[
P(x_1,x_2,\dots,x_n)=\prod_{j =1}^n P(x_j |x_1,\dots,x_{j-1})\,,
\]
where $P(x_j|x_1,\dots,x_{j-1})$ are the conditional probabilities
given the values $x_1,\dots,x_{j-1}$ of $X_1,X_2,\dots,X_{j-1}$.
Let $G$ be a directed acyclic graph. Assume that $G$ represents
the underlying causal structure of $X_1,X_2,\dots,X_{nal}$. An
arrow from $X_j$ to $X_l$ indicates that $X_j$ influences $X_l$
directly (here ``directly'' means that the causal effect is  not
intermediated by another variable $X_m$). Assume that the
variables are ordered in a way that is consistent with $G$, i.e.,
there is no arrow from any $X_j$ to a variable $X_l$ with $l\leq
j$. In case the variables $X_j$ correspond to definite and
different times $t_j$, one may think of this order as the time
order $t_1 < t_2  < \dots <t_n$. Due to the fact that each $X_j$
is only (directly) influenced by its parents (the nodes with an
arrow to $X_j$,) we can give a simpler factorization for $P$ as
follows:
\begin{equation}\label{Faktor}
P(x_1,x_2,\dots,x_n)=\prod_j P (x_j | x_{j;1},x_{j;2},\dots,x_{j;k_j})\,,
\end{equation}
where $X_{j;1},\dots,X_{j;k_j}$ are the $k_j$ parents of $X_j$.
It can be shown  (Pearl, 2000) that this factorization implies the so-called
Markov condition defined as follows.

\begin{definition}
Let $P$ be a joint probability distribution of $n$ random
variables \\$X_1,X_2,\dots,X_n$ and
$G$
a directed acycylic graph  with the variables as nodes.
Then $P$ is said to satisfy the Markov condition relative to $G$
if for each variable $X_j$
the following condition holds:

Given the values of all the parents of  $X_j$, the variable $X_j$ is
statistically independent from the set of those nodes $X_l$ with $l \neq j$
that are
no (direct or indirect)
descendants of $X_j$.

\end{definition}

Here statistical independence of two sets  $\cX:=\{X_1,\dots,X_k\},
\cY:=\{Y_1,\dots,Y_l\}$ of variables
given a third set $\cZ:=\{Z_1,\dots,Z_m\}$ is defined by the condition
\[
P(x_1,x_2,\dots,x_k,y_1,y_2,\dots,y_l |z_1,\dots,z_m)
=P(x_1,\dots,x_k|z_1,\dots,z_m)P(y_1,\dots,y_l|z_1,\dots,z_m)
\]
for all possible assignments of the values $x_i,y_j,z_r$.
Every graph $G$ defines a set of probability distributions:

\begin{definition}
Let $G$ be an arbitrary directed acyclic graph on $n$ nodes labeled
with the random variables $X_1,\dots,X_n$.
Then $\cP_G$ is the set of all joint probability
distributions of $X_1,\dots,X_n$
that satisfy the  Markov condition relative to $G$.
\end{definition}

Given an arbitrary order on $X_1,X_2,\dots,X_n$ we define the {\it
complete acyclic graph} $G_c$  corresponding to the order as the
graph with an arrow from each $X_j$ to each $X_l$ with $l>j$. Note
that every probability measure is Markovian relative to $G_c$.
Hence one can find graphs $G$ such that $P$ is Markovian relative
to $G$ by testing which edges can be removed from $G_c$ without
violating the Markov property. Then a Bayesian network is formally
defined as the pair $(G,P)$ where $G$ is a directed acyclic  graph
with random variables as nodes and $P$ a joint probability measure
satisfying the Markov property relative to $G$.

Of course a graph $G$ does not necessarily coincide with the true
causal structure when the measure is Markovian relative to $G$. We
do not focus on the deep problem of inferring causal structure
from statistics (Pearl, 2000). Here we mentioned the causal point
of view only to emphasize that {\it simple} Bayesian networks may
stem from a {\it simple} causal structure. The goal in this
article is not to infer the causal structure but rather to infer
properties of the probability measure from sparse data.

It is interesting to note that the graph $G$ determines directly
the free
parameters of those probability measures that are
Markovian relative to $G$
since
the probability measure $P$ is
determined once the ``transition probabilities'' $P (x_j |
x_{j;1},x_{j;2},\dots,x_{j;k_j})$ are given. Once we have found a
hypothetical graph with  corresponding transition probabilities
that seems to be in good agreement with the observed data we would
like to judge whether we have really found a good model or whether
the good agreement is rather caused by over-fitting our limited
amount
of data. In (Wocjan and Janzing, 2002) upper bounds on the
required sample size for learning the probabilities of all
$k$-tuples $(x_{j_1},x_{j_2},\dots,x_{j_k})$ with a certain
accuracy and reliability are given. Under the assumption that the
true probability measure is Markovian relative to a simple graph
this will give the joint measure on the $n$ variables up to a
certain accuracy. Here we do not make any prior assumptions on the
underlying joint probabilities. Merely the fact that we have found
a simple model that does explain data shall give the accuracy and
reliability of our guess.

For each node $X_j$, the number of free parameters increases
exponentially with the number $k_j$ of parents of $X_j$. Therefore
it seems to be a reasonable concept to bound the number of  free
parameters of the available probability measures  by considering
graphs with small in-degree (i.e., the maximal number of parents)
in order to avoid over-fitting. However, this is a heuristic
argument. In the context of learning theory, Vapnik (1998) has
argued that a small number of free parameters of the set of
available functions to fit observed data is neither sufficient nor
necessary to avoid over-fitting. He showed the so-called
Vapnik-Chervonenkis (VC) dimension of a set of functions to be
decisive. But we will show that it does make sense from the point
of view of learning theory to consider graphs with small in-degree
since we can derive upper bounds on the VC dimension of the set of
corresponding probability measures.

In Section \ref{Hypotest} we formulate the criterion for judging
whether a hypothetical measure fits well
the observed data and
give a
meaning on what defines a ``good guess'' of
statistical dependencies, i.e. we define a risk functional
quantifying the goodness of fit. In Section \ref{sec:Risc} we
rephrase the concept of VC dimension and explain the general idea
to use it for obtaining reliability bounds when an unknown
function is to be learned. In our case the unknown function is the
joint probability distribution on $n$ random variables. Therefore
we derive in Section \ref{graphVC} bounds on the VC dimension of
sets of joint distributions corresponding to given graphs and
sets of graphs. We show how to obtain reliability bounds for the
estimated distribution. In Section \ref{mini} we show how to apply
structural risk minimization principle in order to learn Baeysian
networks reliably. In Section \ref{Opt} we implement the
minimization as a convex optimization problem.

\section{Selection criterion  for  hypothetical networks}
\label{Hypotest}

Assume the data are given  by $l$ $n$-tuples
\[
  {\bf x}^1,{\bf x}^2,\dots, {\bf x}^l\,.
\]
Using
prior knowledge on the underlying causal structure we
might
prefer a specific graph $G$. It
should have the property that a probability distribution in $\cP_G$
describes the observed data already very well.
In order to find the best
distribution
in $\cP_G$ we use the
following approach (Vapnik, 1998). Consider  the observed relative
frequencies $H({\bf x})$ formally as a probability measure over
the set of possible $n$-tuples $\Omega:=\Omega_1\times
\Omega_2\times \cdots \times \Omega_n$ and minimize the
Kullback-Leibler relative entropy (Cover and Thomas, 1991)
\[
K(\tilde{P}|H):=\sum_{{\bf x}\in \Omega} H({\bf x}) \ln \tilde{P}({\bf x}) -
\sum_{{\bf x} \in \Omega} H({\bf x}) \ln H ({\bf x})\,,
\]
which is equivalent to the minimization of
\begin{equation}\label{eq:Risc}
R_{emp}(\tilde{P}):= \sum_{{\bf x}} H({\bf x}) \ln \tilde{P}({\bf x})=
\frac{1}{l}\sum_{i\leq l} \ln \tilde{P}({\bf x}^i)\,.
\end{equation}
over all $\tilde{P}\in \cP_G$.
By the law of large numbers $R_{emp} (\tilde{P})$ converges to
\[
R(\tilde{P}):=\sum_{{\bf x}\in \Omega} P({\bf x}) \ln
\tilde{P}({\bf x})\,,
\]
where $P$ is the true probability measure on $\Omega$. Note that
$R(\tilde{P})$ yields for  $\tilde{P}=P$
\[
R(P)=\sum_{{\bf x}\in \Omega} P({\bf x}) \ln P({\bf x})=S(P)\,,
\]
i.e., the entropy of $P$. It measures the quality of the
hypothesis concerning the  statistical dependencies since it
measures whether those events that have been predicted to be
rather unlikely by the hypothetical measure do really occur
rarely. Therefore a low value of $R(\tilde{P})$ does  not only
mean that the Kullback Leibler distance between $\tilde{P}$  and
the true measure $P$ is low but it also implies that the entropy
of $P$ is low.
This means that we have found strong  statistical
dependencies. They may, for instance, indicate strong causal
influences among the  variables.
This justifies to consider
$R(\tilde{P})$ as  a criterion that measures whether the
hypothetical measure $\tilde{P}$  is not only good in the sense
that its deviation from $P$ is small but also that we have found a
law with high predictive power.

The minimization above is quite convenient since the factorization
in eq. (\ref{Faktor}) corresponds to a sum of the logarithms of
the conditional probabilities. It is clear that this minimization
does not make sense if $G=G_c$ is the complete acyclic graph for a
given order. Since all measures are in $\cP_{G_c}$ we would
clearly obtain $\tilde{P}=H$ -- a fatal over-fit. This example
shows intuitively that the minimization above leads to
over-fitting when $G$ has too many arrows. In order to consider
this problem from a perspective of statistical learning theory we
rephrase the essential concepts in the next section.

\section{Risk estimation by statistical learning theory}
\label{sec:Risc}

As explained above a major problem in inferring the true
probability measure $P$ from the set of training data is
over-fitting. It is not sufficient that $R_{emp}(\tilde{P}) $
is small, we rather would like to have
$R(\tilde{P})$ small.
Abstractly speaking the problem reads: Given a
family $(f_\alpha)$ of negative functions, consider the data
points ${\bf x}^1{\bf x}^2,\dots,{\bf x}^l$ and choose $f_\alpha$ in
such a way that one can expect with high confidence that
\begin{equation}\label{riskf}
R(f_\alpha):= -\sum_{{\bf x} \in \Omega} f_\alpha ({\bf x})
\end{equation}
is small. In this general setting
the specific form of  $f_\alpha$ is not relevant,
the problem is simply
to choose a function $f_\alpha$ from a family $(f_\alpha)$
such that its expectation value is
maximal.
Statistical learning theory tells the following.
If the family $(f_\alpha)$ is small enough
with respect to a specific measure
we 
can say with high confidence
that for all $\alpha$ the risk
$R(f_\alpha)$ deviates from the empirical risk
\begin{equation}\label{empriskf}
R_{emp} (f_\alpha) := \frac{1}{l}\sum_{j\leq l} f_{\alpha} ({\bf x}^j)
\end{equation}
only by a small amount. (Note the slight abuse of notation. To be
consistent, we should have written $R_{emp} (\ln \tilde{P})$ and
$R(\ln \tilde{P})$ in Section \ref{sec:Risc}. However, this should
not lead to any confusions.) To make this precise we briefly
explain the notion of VC dimension. First we introduce it only for
two-valued functions (``indicator functions'', or
``classifiers'').

\begin{definition}
Let $\Lambda$ be an index set of arbitrary cardinality.
Let  $(f_\alpha)_{\alpha \in \alpha}$ be a set of indicator functions
on $\Omega$. Then the $VC$ dimension of $(f_\alpha)_{\alpha \in \Lambda}$
is the largest natural number $l$ such that there exists
$l$ points ${\bf x}^1,{\bf x}^2,\dots,{\bf x}^l \in \Omega$ with the property
that for every indicator function
$\chi: \{{\bf x}^1,{\bf x}^2,\dots,{\bf x}^l\} \rightarrow \{0,1\}$
there exists a function $f_\alpha$ such that its restriction
to $\{{\bf x}^1,{\bf x}^2,\dots,{\bf x}^l\}$ coincides with $\chi$.
\end{definition}

The definition of VC dimension of arbitrary real-valued functions relies
on the VC dimension of sets of indicator functions:

\begin{definition}
Let $(f_\alpha)_{\alpha\in \Lambda}$ be a family of real-valued functions
on a set $\Omega$. Then the VC dimension of $(f_\alpha)_{\alpha\in \Lambda}$
is the VC dimension of the family of the indicator functions
(``classifiers'')
$(\chi_\mu \circ f_\alpha)_{\mu \in \R, \alpha \in \Lambda}$. Here
$\chi_\mu$ is defined by $\chi_\mu(c)=0$ for $c< \mu$ and
$\chi_\mu (c)=1$ for $c\geq \mu$.
\end{definition}

The following theorem is a corollary from the statements in
(Vapnik, 1998, pp.192, end of Section~5.3):

\begin{theorem}
\label{theo:Lern}
Let $(f_\alpha)$ be a set of measurable
real-valued
functions on $\Omega$ bounded below and above by $A$ and
$B$, respectively. Let $h$ be the VC dimension of the set.

Then for any training data $\bx^1,\bx^2,\dots,\bx^l$ we have
with probability at least $1-\eta$
\[
R(f_\alpha) \le R_{emp} (f_\alpha) +
\phi_{A,B}(l,h,\eta)
\]
with
\begin{equation}\label{confiterm1}
\phi_{A,B}(l,h,\eta):=
(B-A)
\sqrt{ \frac{h(\ln (2l/h) +1)- \ln (\eta/4)+1}{l}}
\end{equation}
for all functions $f_\alpha$.
\end{theorem}

Note that the reliability bound is uniform on the family
$(f_\alpha)$, i.e., with probability $1-\eta$ the difference
between
$R_{emp}(f_\alpha)$ and $R(f_\alpha)$ is {\it for all} $f_\alpha$ bounded
by the second term in eq. (\ref{confiterm1}).

In the following section we will give bounds on the VC dimension
of certain sets of joint probability distributions
of $n$ variables.

\section{The VC dimension associated with a graph or a set of graphs}

\label{graphVC}
The factorization of Markovian joint distributions
in eq. (\ref{Faktor}) is decisive for the
upper bound  on the VC-dimension of $\cP_G$.
Note that the VC-dimension of the  families
\[
(\tilde{P})_{P\in \cP_G}
\]
and
\[
(\ln \tilde{P})_{P\in \cP_G}
\]
coincide.
Note furthermore that $\cP_G$ contains also
all distributions that are Markovian relative to a graph $G'$ whenever
$G'$
was obtained from $G$ by deleting some arrows.
In this sense,  one considers always a set of graphs
when general distributions in $\cP_G$ are considered.

Let $m_j:=|\Omega_j|$ the number of elements of $\Omega_j$.
We find:

\begin{theorem}\label{VCgivenGraph}
Let ${\bf r}_j$ be the indices of the set ${\bf P}_j$ of parents of
$X_j$ with respect
to a given graph $G$.
Then the VC-dimension of $\cP_G$ is at most
\[
N_G:=\sum_{j\leq n} \,\,\,\prod_{i\in ({\bf r}_j \cup \{j\})} m_i\,.
\]
\end{theorem}

\begin{proof}
We show that the logarithms of all probability distributions
in $\cP_G$ can be written as a linear functional in a common
$N_G$ dimensional vector space.
For each set ${\bf j}=\{j_1,\dots,j_k\}\subset \{1,\dots,n\}$ we define
\[
\Omega_{\bf j} := \Omega_{j_1} \times   \Omega_{j_2}\times \dots \times
\Omega_{j_k}\,.
\]
Then we define the vector space $V_j$ as the set of real-valued
functions on
\[
\Omega_{{\bf r}_j \cup\{j\}}\,.
\]
The dimension of $V_j$ is clearly given as
\[
\prod_{i\in ({\bf r}_j \cup \{j\})} m_i
\]
By setting
\[
V:=\oplus_{j\leq n} V_j
\]
we obtain a vector space of dimension $N_G$.
Now we define a vector $f_j\in V_j$ by
\[
f_j(x_j,{\bf p}_j):=\ln P(x_j|{\bf p}_j)\,
\]
and
\[
f:=\oplus_{j\leq n} f_j\,.
\]
For each $n$-tuple ${\bf x}:=(x_1,\dots,x_n)$ we
define a vector
\[
c^{\bf x}:=\oplus_j c_j^{\bf x}   \,\,\, \in V
\]
where
each vector  $c_j^{\bf x}$  is $1$ for the entry that corresponds
to the restriction of ${\bf x}$ to ${\bf p}_j \cup \{j\}$
and $0$ for all the other values.
Then the logarithm of the probability of ${\bf x}$ can be
obtianed by
\[
\ln P(x_1,\dots,x_n)= \langle c^{\bf x}| f\rangle\,.
\]
This shows that the logoarithm can be written as linear functional
in $V$.
The VC dimension of the set of linear functions in
$\R^N$ is $N$ (Vapnik, 1998).
This completes the proof.
\end{proof}

The idea of the proof is quite similar to
the proof of Lemma 2 in (Herrmann and Janzing, 2003).
There we have given an upper bound
on the VC dimension of the set of so-called
$k$-factor log-linear models.
These are probability distributions
with the property that their logarithm can be written as a sum of
functions depending on $k$ variables only.
Here we have considered
a specific factorization coresponding to  a given graph.
This prior knowledge decreases the VC-dimension.

Now we consider the case that no
specific graph is given but
all graphs with a given in-degree are allowed which respect a given order
on the set of random variables.
We find:

\begin{theorem}\label{VCorder}
Let $X_1<X_2<\dots <X_n$ be an ordering
on the set of random random variables.
Let $\hat{\cP}_\Delta$ be the set of all measures that are Markovian
relative to an appropriate graph with in-degree $\Delta$ which is consistent
with the order, i.e., there are only arrows from $X_i$ to $X_j$ for $i<j$.
Then the VC dimension of $\hat{\cP}_\Delta$ is at most
\begin{equation}\label{VCordereq}
N_\Delta:=\sum_{j=1}^n \,\,\,\,
\sum_{{\bf i}} m_j m_{i_1} \dots m_{i_\Delta}\,,
\end{equation}
where the second sum runs over all $\Delta$-subsets
${\bf i}:=\{i_1,i_2,\dots,
i_\Delta\}$ of  $\{1,\dots,j-1\}$.
\end{theorem}

\begin{proof}
We extend the proof of Theorem \ref{VCgivenGraph}.
The definition of each space $V_j$ given there  depends on one particuliar
choice of the parents of $X_j$. Now we have a vector space
$V^{\bf i}_j$ corresponding to each possible choice of
parents of $X_j$ given by one
specific  $\Delta$-subset ${\bf i}$ for each node $j$.
We define
\[
\hat{V}:=\oplus_{j\leq n} \,\,\,
\oplus_{{\bf i}} V^{{\bf i}}_j\,.
\]
One checks easily that $N_\Delta$ is the dimension of $\hat{V}$.
Note furthermore that also the definition of each  $c_j^{\bf x}$ in the proof
of Theorem \ref{VCgivenGraph} depends on one specific choice ${\bf i}$  of
the parents of $X_j$. Hence we obtain now a different vector
$c_j^{\bf x, i}$ for each ${\bf i}$.
In analogy to the proof of Theorem \ref{VCgivenGraph}
we assign a vector $\hat{c}^{\bf x}$ to each $n$-tuple
${\bf x}=(x_1,\dots,x_n)$ by
\[
\hat{c}^{\bf x}:= \oplus_j \oplus_{\bf i} c^{\bf x, i}\,.
\]
Let $\tilde{P}$ be an arbitrary probability measure in $\hat{\cP}_\Delta$.
The proof of Theorem \ref{VCgivenGraph} assigns a vector $f\in V$
to this measure. Note that there is a canonical embedding of the vector
space $V$ introduced in the proof of Theorem \ref{VCgivenGraph} into the
space $\hat{V}$ defined here since each $V_j$ defined
there corresponds to one
specific $V_j^{\bf i}$  here.
With this embedding we have
\[
\ln \tilde{P}(x_1,\dots,x_n) =\langle \hat{c}^{\bf x} | f\rangle\,.
\]
Hence the logarithms of probabilities can be
written as an inner product in a vector space of dimension $N_\Delta$.
This completes the proof.
\end{proof}

If no specific order on the random variables is given
a priori the VC dimension of all graphs with a fixed in-degree
is bounded as follows:

\begin{theorem}\label{VCdelta}
Let $\cP_\Delta$ be the set of probability measures
that are Markovian relative to some graph $G$ with in-degree
$\Delta$.
Then the VC dimension of $\cP_\Delta$ is at most
\begin{equation}\label{VCdeltaeq}
N_\Delta:= \sum_{{\bf j}} \prod_{i \in {\bf j}}  m_i\,,
\end{equation}
where the sum runs over all $(\Delta+1)$-subsets ${\bf j}$
of $\{1,2,\dots,n\}$.
\end{theorem}

The proof follows from the observation that each $P\in \cP_\Delta$
is a $(\Delta+1)$-factor log-linear model, i.e., a probability distribution
with the property that its logarithm can be written as a sum of
functions each depending on at most $\Delta+1$ variables.
Then the bound of Lemma 2 in (Herrmann and Janzing, 2003)
applies.

The following corollary from Theorem \ref{theo:Lern}
shows explicitly how to use the bounds on the VC-dimensions
in order to get reliability bounds on the estimated risk functional.
Note that it is therefore necessary to restrict one's attention to sets of
probability measures which are bounded below.
Explicitly, we define:
Let $\cP^\lambda$ for each $\lambda>0$
be the set of joint distributions $\tilde{P}$  with
the property that
\[
\tilde{P}(x_1,x_2,\dots,x_n) \geq \lambda
\]
for all $n$-tuples in $\Omega$.
Note that we do not assume that the true probability measure $P$
 satisfies this
requirement. Only the hypothetical measure $\tilde{P}$  has to be
bounded. Then the bounds $A$ and $B$ in Theorem \ref{theo:Lern}
are $0$ and $-\ln \lambda$, respectively. We conclude:

\begin{corollary}\label{Cor}
Let $\cP \subset \cP^\lambda$ be a set of joint distributions with
VC-dimension $h$. Then for any training data
$\bx^1,\bx^2,\dots,\bx^l$ we have with probability at least
$1-\eta$
\[
R(\tilde{P}) \le R_{emp} (\tilde{P}) +
\phi_\lambda (l,h,\eta)
\]
with
\begin{equation}\label{confiterm}
\phi_\lambda(l,h,\eta):=
(-\ln \lambda)
\sqrt{ \frac{h(\ln (2l/h) +1)- \ln (\eta/4)+1}{l}}
\end{equation}
uniformly for all $\tilde{P} \in \cP$.
\end{corollary}

Setting $\cP:=\cP^\lambda_G$ we obtain reliability bounds on the
estimated probability measure provided that the graph $G$ has been chosen
{\it in advance}.
With $\cP:=\cP^\lambda_\Delta$ we obtain reliability bounds if
the hypothetical measures are restricted to those that factorize
to a ``simple'' graph (in the sense of small in-degree).

However, the prior restriction to a specific  $\lambda$ and
a specific graph or to graphs with small in-degree
is not acceptable.
An appropriate way to learn Bayesian networks  should also consider
complex graphs provided that sufficiently large sampling
strongly indicate a more complicated dependency among the variables.
Similarly, one should not {\it a priori}
exclude probabilities that are smaller
than a specific value
$\lambda$. For large sample size data may give strong evidence that
some probabilities are indeed small.
On the other hand, the estimation in Corollary \ref{Cor} seems to require
prior restrictions.

This problem is solved by
structural risk minimization principle (Vapnik, 1995 and 1998) in
statistical learning theory.
It uses a hierarchy of increasing sets of hypothetical functions.
Then a  function $g$
from a larger set
is only
preferred compared to a function  $f$ from a smaller set if not only
$R_{emp}(g)< R_{emp}(f)$ but also the bound on $R(g)$ is smaller
than the bound on $R(f)$. We explain this principle in the
following section.

Now we briefly summarize the estimations for the VC dimension of
some interesting set of graphs.
Here we assume that $l$ is the maximum over all values $m_j$.

\begin{itemize}

\item For the VC dimension of a
{\bf given graph $G$} with in-degree $\Delta$ we have
\[
h\leq n \,l^{\Delta +1}\,.
\]
This follows from
Theorem \ref{VCgivenGraph} since
\[
\prod_{i\in ({\bf r}_j \cup \{j\})} m_i  \leq  l^{\Delta +1}\,.
\]

\item For the VC dimension of all graphs with in-degree $\Delta$ which respect
a given order on the nodes we have
\[
h\leq   l^{\Delta +1} \sum_{j=1}^n {j-1 \choose \Delta }\,.
\]
This is due to Theorem \ref{VCorder} since the second sum in
eq. (\ref{VCordereq}) runs over
\[
{j-1 \choose \Delta}
\]
terms.

\item For the VC dimension of the set of all graphs
with in-degree $\Delta$ we have
\[
h\leq   {n \choose \Delta+1} l^{\Delta +1}\,.
\]
This follows from Theorem \ref{VCdelta}
since the sum in eq. (\ref{VCdeltaeq}) runs over
\[
{n \choose \Delta+1}
\]
terms.
\end{itemize}

This seems to suggest that in general a small in-degree
reduces the VC dimension considerably whereas prior knowledge
on the causal order is less relevant.

\section{Structural risk minimization}

\label{mini}

Before we apply structural risk minimization to the problem of learning
probabilities we briefly sketch the general idea.
Consider the case that an arbitrary function on a set $\Omega$
is to be learned.
Define a sequence $(F_k)_{k\in N}$ of  families $F_k$ of functions.
The idea is that the sequence defines a hirarchy of
more and more complex families of functions and the less complex ones are  a
priori preferred. Let $(p_k)_{k\in \N}$ be any sequence of
non-negative numbers with $\sum_k p_k=1$. These values express to
what extent one tends to prefer functions from $F_k$ with lower
$k$. Let $h_k$ be the VC-dimension of $F_k$. Then one has with
probability $1-\eta$ that for each function $f\in \bigcup_k F_k$
\begin{equation}\label{Rel}
R(f) \leq R_{emp}(f) + \phi (h_k,l,p_k \eta)\,,
\end{equation}
where $\phi$ is the confidence term in eq. (\ref{confiterm1}).
This is a standard union bound argument (see e.g. Herbrich, 2002).
Note that the sequence on $p_k$ may be chosen in such way that
it expresses prior probabilities to the choice of a certain class $F_k$.
But it should be emphasized that the reliability bound in eq. (\ref{Rel})
does not rely on this interpretation.

Here we define a  hirarchy of probability measures which takes into account
two aspects  of a measure: We prefer  measures which are Markovian
relative to a simple graph and measures with high cut-off value $\lambda$.
Let $(\lambda_m)_{m\in \N}$ be a sequence of positive values
converging to zero. Let $\cP^{\lambda_m}$ be the set of probability measures
bounded from below by $\lambda_m$. Let $\cP_1,\dots,\cP_r$ be
$r$ sets of probability measures. They may, for instance, correspond to
an enumeration of
all directed acyclic graphs on $n$ nodes. They may also
correspond to graphs with in-degree $1,2,\dots,r$.
Then we prefer probability measures in $\cP_k \cap \cP^{\lambda_m}$ for
small $m$ and small $k$.
We may express this by defgining probabilities
$q_{k,m}$ which are decreasing in $k$ and $m$.
In analogy to the bound above we obtain:

\begin{theorem}
Let $(\cP_k)_{k\in K}$   with $K=\N$ or $K=\{1,\dots,r\}$ an arbitrary set
of
families of
joint distributions on the $n$ random variables.

Let $(q_{k,m})_{k,m}$
define an arbitrary probability measure on $K\times \N$.
Let $h_k$ be the VC-dimension of $\cP_{k}$. Then we know
with probability $1-\eta$
for all $k\leq r, m\in \N$
and all $\tilde{P} \in \cP_{k} \cap \cP^{\lambda_m}$
\[
R (\tilde{P}) \leq  R_{emp}(\tilde{P}) + \phi_\lambda (h_k,l,q_{k,m} \eta)
\]
holds, with
\[
\phi_{\lambda_m}
(h_k,l, q_{k,m} \eta) := -\ln \lambda_m 
\sqrt{ \frac{h_k(\ln (2l/h_k) +1)- \ln (q_{k,m}\eta/4)+1}{l}}
\]
\end{theorem}

The structural risk minimization principle  works as follows.
For a given number $k, m$ choose $\tilde{P}_{k,m}
\in \cP^{\lambda_m} \cap \cP_k$
such that
$R_{emp}(\tilde{P}_{k,m})$ is minimal.
Then choose $k,m$ such that
\[
R_{emp} (\tilde{P}_{k,\lambda}) +
\phi_{\lambda_m} ( l,h, q_{k,m} \eta)
\]
is minimal.

The following example gives an idea how to apply this principle.
Given $n$ binary variables.
Then our upper bound on the VC dimension
of the set of graphs with in-degree $\Delta$ is
\[
h_\Delta \leq n \,2^{\Delta +1}\,.
\]

Let $\cP_k$ be the set of joint distributions which are Markovian
relative to a graph with in-degree $k$. Set furthermore
$\lambda_m=2^{-m}$ and choose the prior probability measure
on $\N\times \N$ as
\[
q_{k,m}:=2^{-k-m}\,.
\]
For $\tilde{P}\in \cP_k \cap \cP^{\lambda_m}$ we obtain
\[
\phi_{\lambda_m} (l, h_k, q_{k,m}\eta) =
m \ln 2 \sqrt{ \frac{n2^{k+1}(\ln (2l/(n2^{k+1})) +1)- \ln (\eta/4)
+(k+m)\ln 2+1}{l}}
\]
The confidence term grows exponentially in $k$, with $O(\sqrt{n})$  and
with $O(m^{3/2})$ whenever the other parameters are fixed.
Hence the required sample size grows quickly with the in-degree, whereas
the number of random variables is less decisive.
Also the  cut-off value $\lambda$
of the probabilities
is less decisive since the required sample size grows only
with $O(m^{3/2})$ although we have defined the cut-off values  $\lambda_m$
in such a way that they  decrease
exponentially in $m$.

\section{Convex Optimization  for Bayesian networks}
\label{Opt} The number of directed acyclic graphs with constant
in-degree $\Delta$ and $n$ nodes increases polynomially in $n$.
Therefore it is realistic to assume that for all graphs with small
in-degree (``sparse graphs'') the optimization can be carried out
for each graph. Hence we may restrict our attention to finding the
optimal probability measure that is Markovian relative to a given
graph $G$ and bounded by a given value $\lambda$ from below. Let
$V_j$ be defined as in the proof of Theorem \ref{VCgivenGraph},
i.e., the set of real-valued functions on
\[
\Omega_{{\bf r}_j \cup\{j\}}\,.
\]
Let ${\bf x}^i$ be the $i$-th observed $n$-tuple.
Let ${\bf x}^i|_{{\bf r}_j \cup \{j\}}$
its restriction to the variable $X_j$ and all its parents.
Then
the task is to find a vector
\[
f=\oplus_j f_j \in  \,\,\oplus_j V_j =V
\]
that minimizes
\[
R_{emp} (f):= \frac{1}{l} \sum_{i\leq l} \sum_{j\leq n}
f_j ({\bf x}^i|_{{\bf r}_j \cup \{j\}})
\]
subject to the following constraints:

\begin{enumerate}

\item
For each $j$ the sum of the conditional probabilities $P(x_j|\omega)$
over all $x_j \in \Omega_j$ has to be $1$ for all $\Delta$-tuples
$\omega \in \Omega_{{\bf p}_j}$.
Formally this means
\begin{equation}\label{Norm}
Z_\omega (f_j):= \exp( \sum_{x_j \in \Omega_j} f_j(x_j,\omega))=1
\end{equation}
for $\omega \in \Omega_{{\bf p}_j}$.

\item
No probability $P(x_1,\dots,x_n)$ is less than $\lambda$.
We can achieve this by  stating the stronger constraint
that no transition probability $P(x_j|\omega)$ is less than
$\lambda^{1/n}$.
This is equivalent to
\[
G_{\omega,j,x_j} (f) :=f_j(x_j,\omega) \geq \frac{1}{n}\ln \lambda\,.
\]

\end{enumerate}

The optimization is rather similar to that one in (Herrmann and Janzing, 2003)
with the decisive difference that the normalization
can be performed for each node separately here
whereas the normalization condition for the joint measure on $n$
variables involves a sum over all possible $n$-tuples, i.e.,
a number growing exponentially in $n$.
Here the computational complexity grows only polynomially in
$n$ for constant $k$. The number of constraints grows linearly in $n$
but exponentially in $k$. The number of terms in the sum
(\ref{Norm}) grows also exponentially in $k$. But since we assume that
$k$ is small we consider the optimization as computationally tractable.
Due to the convexity of the constraints (see Herrmann and Janzing, 2003)
it is a usual linear programming problem that can be efficiently
solved (Pallaschke and Rolewicz, 1997).

\section{Conclusions}

%

We have presented a method for estimating the joint distribution
of a large number of random variables from sparse data. 
The statistical dependencies among the variables
are explained by Bayesian networks such that networks with simple
graphs are preferred.
We provide
reliability bounds  without restricting the
set of joint distribution under consideration. We have shown
that the set of probability measures that are markovian relative
to simple graphs have low VC-dimension.  This guarantees reliable
estimation in the sense of statistical learning theory whenever
the observed data is explained well  by those ``simple measures''. 
If no simple Bayesian network fits the data the method 
does not allow reliable estimation.
Furthermore we have shown that finding the optimal
distribution within a class of distributions (markovian relative to a
given graph) is a convex optimization problem. Since the number of
simple graphs is not too large, the whole estimation can be
performed efficiently.

\end{document}